# Exploring CNN-based models for image's aesthetic score prediction with using ensemble

Ying Dai

Iwate Prefectural University, Takizawa, Japan

dai@iwate-pu.ac.jp

**Abstract:**

In this paper, we proposed a framework of constructing two types of the automatic image aesthetics assessment (IAA) models with different CNN architectures and improving the performance of the image's aesthetic score (AS) prediction by the ensemble. Moreover, the attention regions of the models to the images are extracted to analyze the consistency with the subjects in the images. The experimental results verify that the proposed method is effective for improving the AS prediction. Moreover, it is found that the AS classification models trained on XiheAA dataset [25] seem to learn the latent photography principles, although it can't be said that they learn the aesthetic sense.

## 1. Introduction

Automatic image aesthetics assessment (IAA) can be applied to a variety of tasks, such as image recommendation, image retrieval, photo management, and product design (cooking). In [1], the authors give an experimental survey about this field's research. In this paper, besides the discussion of the main contributions of the reviewed approaches, the authors systematically evaluate deep leaning settings that are useful for developing a robust deep model for aesthetic scoring. Early efforts of IAA focus on extracting designed hand-crafted features according to the known photographic principles, for example, the rule of thirds, color harmony, and global image layout [2-5]. With the advance of convolutional neural network (CNN), recent methods aim to map image aesthetics to different types of tasks using CNNs, majorly including high/low quality classification, aesthetic score prediction and their distribution [6-11]. Although some achievements have obtained, the state-of-art research involves the attention mechanism the and layout-aware graph convolutional network in IAA, so as to improve the performance of the aesthetic score prediction, and so on.

In [12], a multi-patch aggregation method for image aesthetic assessment with preserving the original aspect ratio is proposed. In this method, the goal is achieved by resorting to an attention-based mechanism that adaptively adjusts the weight of each patch of the image. In [13], the authors propose a gated peripheral–foveal convolutional neural net-work. It is a double-subnet neural network. The former aims to encode the holistic in-formation and provide the attended regions. The latter aims to extract fine-grained features on these key regions. Then, a gated information fusion network is employed for the image aesthetic prediction. In [14], the authors propose a novel multimodal recurrent attention CNN, which incorporates the visual information with the text information. This method employs the recurrent attention network to focus on some key regions to extract visual features. In [29, 30], the contributions of different regions at object level to aesthetics are adaptively predicted. However, it has been validated that feeding the weighted key regions to CNN to train the IAA model degrades the performance of prediction according to our preliminary experiments, because the aesthetic assessment is influenced by holistic information in the image. Weakening some regions results in the information degradation for aesthetic assessment.

In [31], a hierarchical layout-aware graph convolutional network is involved to capture layout information for unified IAA. However, although there is a strong correlation between image layouts and perceived image quality, the image layout is neither the sufficient condition nor the necessary condition to determine the image's aesthetic quality. In fact, several typical failure cases presented in [31] confirm the above statement. Figure 5 in the paper shows several failure cases. Some pictures appear the good lay-outs that seem to meet the rule-of-thirds and are predicted to have a high rating. However, the ground truths (GT) of these images are of low ratings. A picture seems not to meet the photography composition principles and is assigned to a low rating. However, its GT is of high rating.

Generally, modeling IAA is supervised learning. Most of the research utilize the labeling data of the images regarding aesthetics in the public photo dataset, such as CUHK-PQ [1] or AVA [28], to train the model. However, these aesthetic data are almost labeled by the amateurs. Whether the labeling data embody the latent principles of aesthetics is not clear. So, whether the IAA models trained on these datasets are significant is also unclear. To make the labelled data embody the photo's aesthetic principles, the author in [25] aims to establish a photo dataset called XiheAA which are scored by an experienced photographer, because it is assumed that the experienced photographers should have the higher ability of reflecting the latent principles of aesthetics when they assess the photos. These labelled images are used to train the IAA model. However, the IAA exhibit a highly-skewed score distribution. in order to solve the imbalance issue in aesthetic assessment, in this paper, the author proposes a method of repetitive self-revised learning (RSRL) to retrain the CNN-based aesthetic score prediction model repetitively by transfer learning, so as to improve the performance of imbalance classification caused by the overconcentration distribution of the scores. Moreover, in [32], the author focuses on the issue of CNN-based RSRL to explore suitable metrics for Establishing an Optimal Model of IAA. Further, the learned feature maps of the model are utilized to define the first fixation perspective (FFP) and the assessment interest region (AIR), so as to analyze whether the aesthetics features are learned by the optimal model. Although RSRL shows the effectiveness on the imbalance classification by several experiments, how to construct an aesthetic score prediction model which really embodies the aesthetic principles on IAA is not involved.

In photography, it is known that two important elements of assessing a photograph are the subject and the holistic composition. One standard for a good photograph is that the image should be achieve attention-subject consistency. Inspired by the above knowledge, we propose a framework of constructing two types of IAA models with different CNN architectures and improving the performance of the image's AS prediction by the ensemble, and analyzing the consistency of the subject with the attention regions of the models. The contributions of the paper are summarized as follows.

- Besides fine-tuning the pretrained models, a new CNN architecture which could embody the holistic composition of the image is designed. Based on this architecture, the models with different architectural parameters are trained on XiheAA dataset [25] to predict the image's aesthetic score.
- The performances of the above models are evaluated, and an ensemble method of aggregating two models is proposed to improve the performance of the AS prediction.
- The feature maps of the models regarding the images are analyzed. It is found that the attention regions of the models are often consistent with the subjects of the images, and follow the simple photography composition guidelines, such as visual balance, and rule of thirds, if they are predicted to have the high aesthetic scores, otherwise the opposite, whether or not the correct predictions are made. It is indicated that the models trained on XiheAA seem to learn the latent photography composition principles, but it cannot be said that they learned the aesthetic sense.

## 2. Related work

**Image Aesthetics Assessment (IAA)**　Besides the research mentioned in the Section Introduction, the other main-stream research on IAA is as the following.

In [15], the authors propose a unified algorithm to solve the three problems of image aesthetic assessment, score regression, binary classification, and personalized aesthetics based on pairwise comparison. The model for personalized regression is trained on the FLICKERAES dataset [16]. However, the ground truth score was set to the mean of five workers' scores. Accordingly, whether the predicted score embodies the inherently personal aesthetics is not clear.

On the other hand, some researchers aim at extracting and analyzing the aesthetic features to find the relation with the aesthetic assessment. In [17], the paper presents an in-depth analysis of the deep models and the learned features for image aesthetic assessment in various viewpoints. In particular, the analysis is based on transfer learning among image classification and aesthetics classifications. The authors find that the learned features for aesthetic classification are largely different for those for image classification; i.e., the former accounts for color and overall harmony, while the latter focus-es on texture and local information. However, whether this finding is universal needs to be validated further. In [18], besides extracting deep CNN features, five algorithms for handcrafted extracting aesthetic feature maps are proposed, which are used to ex-tract feature maps of the brightness, color harmony, rule of thirds, shallow depth of field, and motion blur of the image. Then, a novel feature fusion layer is designed to fuse aesthetic features and CNN features to improve the aesthetic assessment. However, the experimental result shows that the fusion only improves the accuracy of 1.5% over no-fusion. Accordingly, whether it is necessary to incorporate the inefficiently hand-crafted aesthetic features with deep CNN features needs to be investigated.

Recently, the fusion technologies are focused on improving the accuracy of the aesthetics assessment. In [19], the authors introduce a novel, deep learning-based architecture that relies on the decision fusion of multiple image quality scores coming from different types of convolutional neural networks. The experimental results show that the proposed method can effectively estimate perceptual image quality on four large IQA benchmark databases. In [20], the authors propose an aesthetic assessment method, which is based on multi-stream and multi-task convolutional neural networks (CNNs); this method extracts global features and saliency features from an input image. These provide higher-level visual information such as the quality of the photo subject and the subject–background relationship.

On the other hand, applying image aesthetics assessment to the design field becomes a hot topic in the resent years [21-24]. For example, in [21], the impact of cover image aesthetics on content reading willingness is analyzed. In [22], the food images are assessed by learning its visual aesthetics. In [23], the abstract images are generated by using correction structure with the aesthetics.

**XiheAA dataset [25]** This dataset contains 3100 photos aesthetically scored by an experienced photographer. Those photos were taken by the students of the photographer's class. The scores range from 2 to 9. So, the number of classes N=8. The distribution of the scores is shown in Fig. 1.

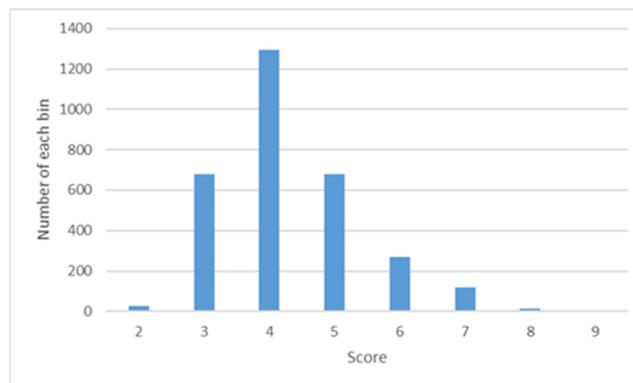

Fig. 1 Score distribution of XiheaAA dataset

**RSRL [25]** The approach of CNN-based RSRL is to drop out the low likelihood samples of the majority classes of scores repetitively, so as to ameliorate the invasion of these samples to the minority classes and prevent the loss of the samples with discriminative features in the majority classes. In this process, the previous model is re-trained by transfer learning again and again. Accordingly, many re-trained models are generated with RSRL. Then, the optimal model is determined among these models based on the F-measures.

**FFP and AIR [32]**    As it is known, people usually focus on the most salient object which is considered as the first fixation perspective (FFP), and the relations with other elements which are considered to be the assessment interest region (AIR), when they enjoy photos. For a CNN-based IAA model, it is supposed that the most activated feature map should be related to the FFP of the image, and the sum of feature maps should be related to the AIR.   So, the FFP is obtained by the most activated feature map of the last convolutional layer, and the AIR is calculated by the sum of feature maps of that.

## 3.  Methodology
### 3.1 Overview

An overview of our proposed framework of constructing two types of IAA models with different CNN architectures and improving the performance of the image's AS prediction by the ensemble is shown in Fig. 2.

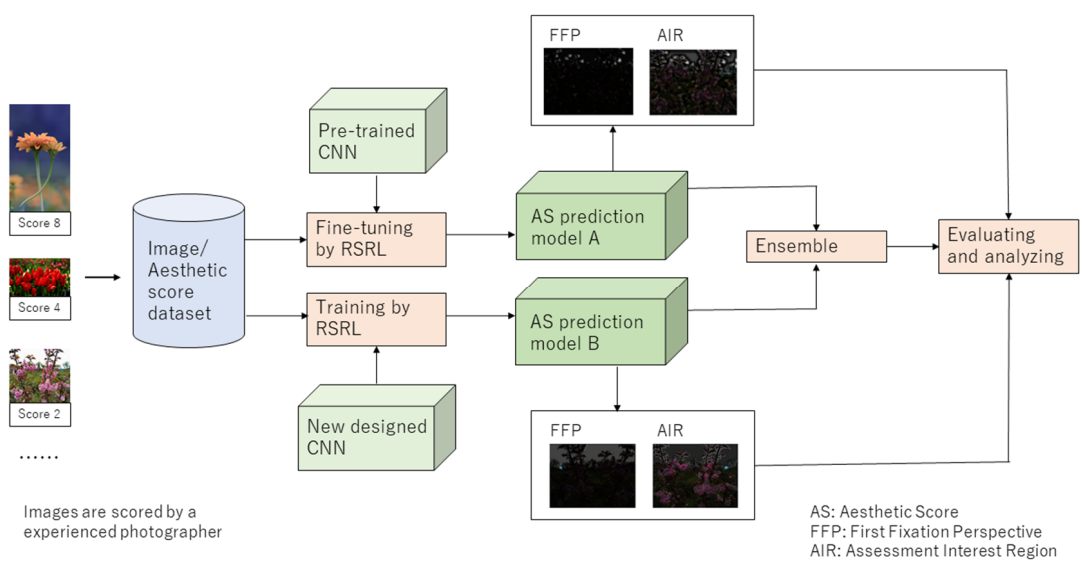

Fig. 2 An overview of the proposed method

For this framework, two types of CNN-based models are trained by RSRL on XiheAA dataset. The model A is expected to extract the subject of the image for predicting the image's aesthetic score (AS), and the model B is expected to extract the holistic composition for the prediction. Because the pretrained models, such as alexNet, resNet18, and efficientNetB0, are trained on ImageNet following the general classification task, and therefore cover a wide range of contents, it is suitable to use such models to construct the model A by transfer learning on XiheAA dataset, while the number of the classes are changed from 1000 to 8, because the scores rated on XiheAA is in the range of [2,9]. On the other hand, because the XiheAA dataset is rated by an experienced photographer, it is considered that a new designed CNN which is trained on it could construct the model B that reflects the holistic composition of the images. Moreover, the ensemble of model A and model B is applied to improve the performance of the prediction. Next, the FFPs and AIRs of model A and model B are computed to analyze the consistency of the attention regions of the models with the photography principles. In the following, the architectures of the new designed CNN, the method of the ensemble, and the FFPs and AIRs of the images regarding model A and model B are explained in details.

**3.2 New designed CNN architecture**

Inspired by the attention mechanism and the architecture of EffcientNetB0 [33], a new CNN architecture is designed for the image's AS classification. The architecture is shown in Fig.3.

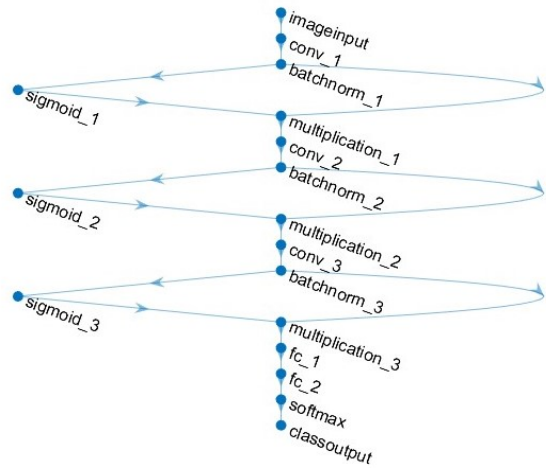

Fig.3 A new designed architecture

This architecture consists of three convolutional blocks (CB) and two full connection layers, fc_1 and fc_2. The elements of CB include convolutional layer, batchnom layer that execute the batch normalization, sigmoid layer, and multiplication layer that execute the element-wise multiplication of the elements in the batchnom layer and the sigmoid layer. By adjusting the parameters of this architecture, four kinks of networks are constructed, which all called model B1, B2, B3, and B4. Especially, for the model B3 and model B4, the resolution of the input images is set up to 192x192. This is for the purpose to make the resolution of the input to the fc_1 is 6x6, which can embody the composition regarding the rule of thirds. The details of these networks are shown in Table 1, 2 3, and 4. The conv1x1 means that the size of the filter of the convolutional layer is 1x1, and the conv3x3 means the size of the filter is 3x3.

Table 1 Model B1

| Stage | Operator | Resolution | Channels |
|---|---|---|---|
| 1 | CB1, conv1x1 | 227x227 | 128 |
| 2 | CB2, conv1x1 | 28x28 | 96 |
| 3 | CB3, conv1x1 | 7x7 | 96 |
| 4 | fc_1 | 7x7 | 36 |
| 5 | fc_2 | 1x36 | 8 |

Table 2 Model B2

| Stage | Operator | Resolution | Channels |
|---|---|---|---|
| 1 | CB1, conv1x1 | 227x227 | 128 |
| 2 | CB2, conv1x1 | 28x28 | 96 |
| 3 | CB3, conv3x3 | 7x7 | 96 |
| 4 | fc_1 | 7x7 | 36 |
| 5 | fc_2 | 1x36 | 8 |

Table 3 Model B3

| Stage | Operator | Resolution | Channels |
|---|---|---|---|
| 1 | CB1, conv1x1 | 192x192 | 128 |
| 2 | CB2, conv1x1 | 24x24 | 96 |
| 3 | CB3, conv1x1 | 6x6 | 96 |
| 4 | fc_1 | 6x6 | 36 |
| 5 | fc_2 | 1x36 | 8 |

Table 4 Model B4

| Stage | Operator | Resolution | Channels |
|---|---|---|---|
| 1 | CB1, conv1x1 | 192x192 | 128 |
| 2 | CB2, conv1x1 | 24x24 | 96 |
| 3 | CB3, conv3x3 | 6x6 | 96 |
| 4 | fc_1 | 6x6 | 36 |
| 5 | fc_2 | 1x36 | 8 |

### 3.3 Ensemble

The ensemble of model A type and model B type is used to improve the performance of the AS prediction. In details, let the probability of assigning an image to an AS class by a classification model is denoted as $p_s^{model}$. Where, *model* indicates the model type, and *s* does the AS class. So, the ensemble of model A and model B is calculated by the equation as below.

$$p_s^{ensemble} = w_1 p_s^{model\ A} + w_2 p_s^{model\ B} \qquad (1)$$

Where, $w_1$ and $w_2$ are the weights of $P_s^{model\ A}$ and $P_s^{model\ B}$, respectively. Then, the predicted score after the ensemble is obtained by the following expression.

$$\text{score} = \underset{s}{\arg\max}\{p_s^{ensemble}, s \in [2, 9]\} \qquad (2)$$

### 3.4 FFP and AIR

FFP and AIR of a model A or a model B can be calculated by [32]. An example of the image regarding its FFP and AIR obtained by the model A and the model B is shown in Fig. 4.

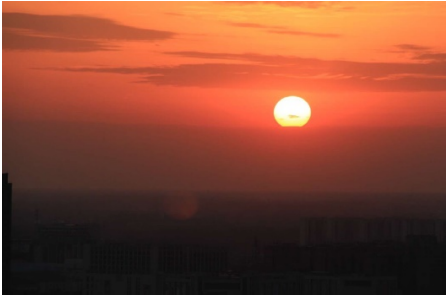

(a) Original image

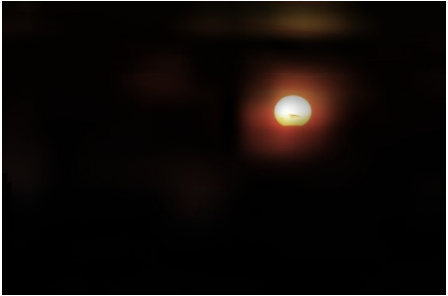 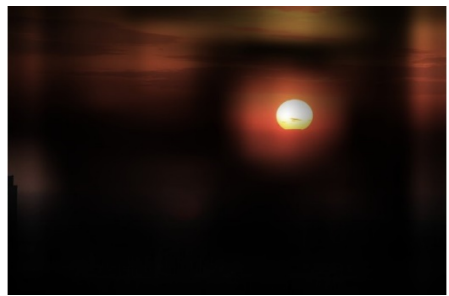

(b) FFP by model A  (c) AIR by model A

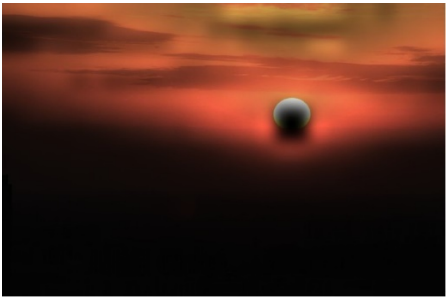 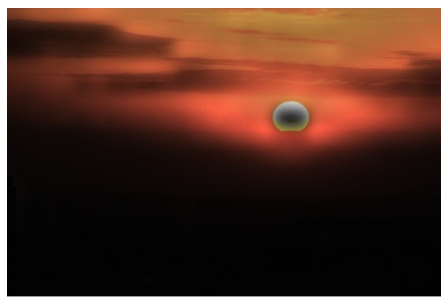

(e) FFP by model B  (f) AIR by model B

Fig. 4 Examples of FFP and AIR

The model A is trained by transfer learning based on the resNet18, and the model B is trained based on the architecture of model B3. The score assigned by model A is 5, and the score assigned by model B is 4. Meanwhile, the ground truth of this image in XiheAA dataset is 4. From the results, we can see that the FFP and the AIR extracted by the model A is the sun that is the subject of the image, and its surrounding, while these extracted by the model B is the red cloud around the sun that reflects the composition of the image. The sun is in the position that seems to meet the rule of thirds, so that the score assigned by the model A is higher than the ground truth. While, the layout of the cloud is dull, so that the score assigned by the model B is 4 that is same as the ground truth. This observation is consistent with the above expectation that the model A extracts the subject of the image for predicting the image's AS, and the model B extracts the holistic composition for the prediction.

## 4  Experiments and analysis

Metrices of precision, recall, F1, and accuracy are sued to evaluate the performance of the models.

### 4.1 Experimental results of single models

The F1values of the AS classes for the various single models on the test dataset of XiheAA are shown in Fig. 5. Ares, Aeff, and Aalex indicate the model A type trained by the transfer learning based on the pretrained models resNet18, efficientNetB0 and alexNet, respectively. B1, B2, B3, and B4 indicate the model B type with the architectures of model B1, model B2, model B3. And model B4, respectively. The assigned scores are in the range of [2,7].

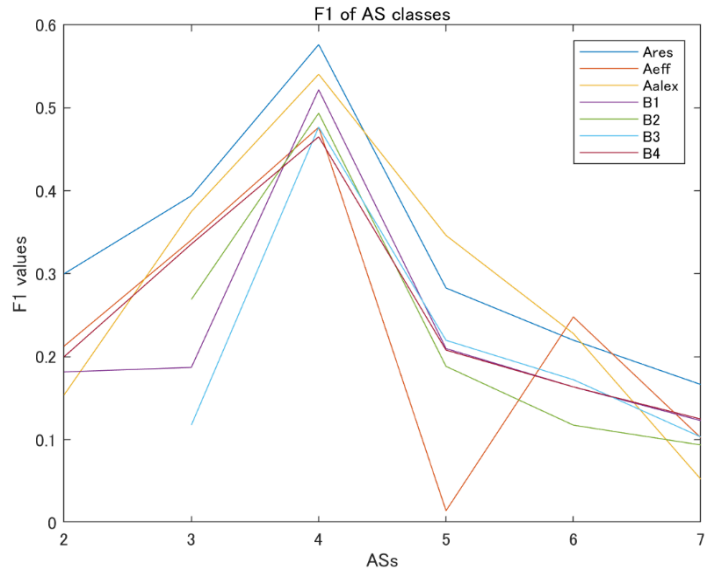

Fig. 5 F1 values for various single models

For the model A group, the model Ares outperforms the other models. For the model B group, the model B1, B3 and B4 outperforms the model B2 for predicting the highly rated images. However, the model B4 outperform the others, especially the model B2 and B3, for predicting the lowly rated images.

From the above observation, it seems that the model B4 with the architecture that the input size is 192x192 and the filter size of the last convolutional layer is 3x3 is most suitable for the AS classification. However, for the model A group, the model Ares trained by the transfer learning based on the resNet18 is best for the AS classification.

Fig. 6 shows the average values of precision, recall, and F1 of the various models to all the AS classes on the test dataset of XiheAA.

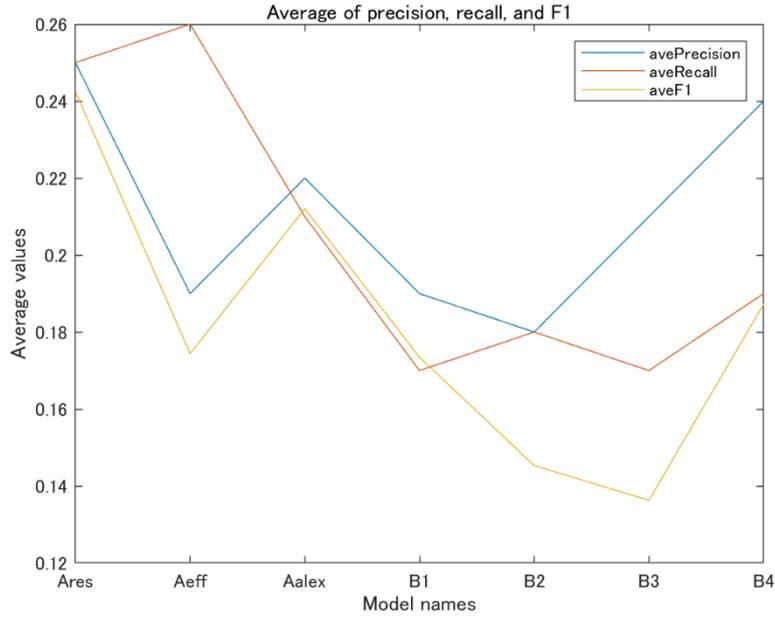

Fig. 6 Average of precision, recall, and F1

For the precision, the average values of the model Ares and the model B4 are comparatively high. For the recall, those of the model Ares, the model Aeff and the model B4 is comparatively high. For the F1 which reflects the comprehensive performance of the classification, the average values of the model Ares, model Aalex and the model B4 are higher. Accordingly, we can see that the model Ares in the A group and the model B4 in the B group have the best performance for the AS classification.

### 4.2 Ensemble

Based on the above observation, the model Ares and the model B4 are used for the ensemble based on equation (1). Then, the AS of the image is predicted by the equation (2). Fig.7 shows the average F1 values of the AS classes on the test dataset of XiheAA with adjusting the weights $w_1$ and $w_2$.

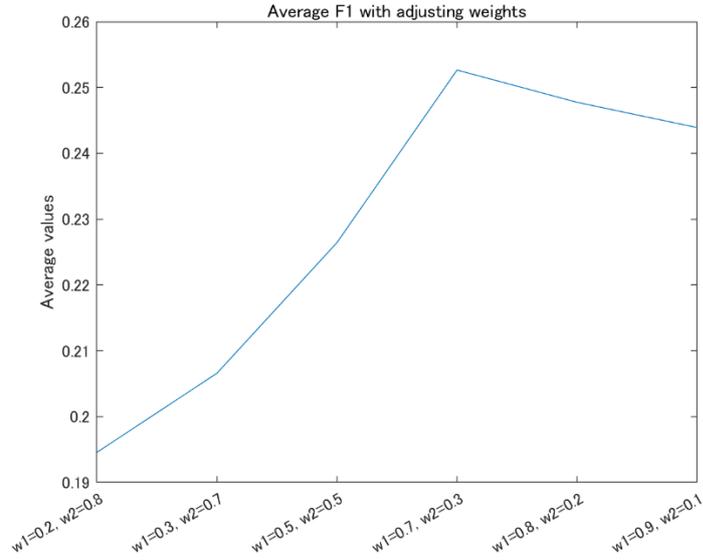

Fig.7 Average F1 with adjusting weights

It is obvious that the average F1 is maximal when $w_1 = 0.7$ and $w_2 = 0.3$. The value of this is 0.253. However, the average value of F1 of the model Ares is 0.24, and that of the model B4 is 0.19. Accordingly, the average F1 of the ensemble improves 5.4% over the model Ares, and 33.1% over the model B4. Moreover, we can see that the weight of the Ares is larger than that of the B4 while the average F1 is maximal. That is, the influence of the model A is stronger than the model B in the AS prediction, although the ensemble can improve the above performance.

### 4.3 Experimental results on CUHK-PQ dataset [1]

The CUHK-PQ dataset [1] is used for the out-of-distribution validation. The CUHK-PQ dataset contains 10,524 high-quality images and 19,166 low-quality images. So, the images predicted to have the score of less than 5 are assigned to the low class, while the others are assigned to the high class. Table 5 shows the accuracy and the averages of precision, recall, and F1for the various models and ensembles.

Table 5 The accuracy and the averages of precision, recall, and F1

| Model | accuracy | avePrecision | aveRecall | aveF1 |
|---|---|---|---|---|
| Ares | 0.650 | 0.628 | 0.636 | 0.630 |
| Aeff | 0.633 | 0.641 | 0.654 | 0.628 |
| Aalex | 0.619 | 0.578 | 0.576 | 0.576 |
| B1 | 0.615 | 0.576 | 0.575 | 0.575 |
| B2 | 0.577 | 0.546 | 0.548 | 0.546 |
| B3 | 0.617 | 0.576 | 0.573 | 0.573 |
| B4 | 0.618 | 0.576 | 0.572 | 0.573 |
| 0.7Ares+0.3B1 | 0.673 | 0.634 | 0.617 | 0.621 |
| 0.7Ares+0.3B3 | 0.674 | 0.636 | 0.617 | 0.621 |
| 0.7Ares+0.3B4 | 0.674 | 0.636 | 0.617 | 0.621 |

From the Table 5, we can see that the performances of the model Ares and Aeff outperform the model Aalex for the model A group. For the model B group, the performances of the model B1, B3, and B4 are almost same, and outperform the model B2. For the ensemble, 0.7Ares+0.3B1 indicates that the model Ares and the model B1 are used, and the weights of them are 0.7 and 0.3, respectively. 0.7Ares+0.3B3 and 0.7Ares+0.3B4 are the analogized ones. From the results of the ensembles, we can see that the performances of the above ensembles are almost same. The accuracies are higher than the all single models. The improved rates are in the range of [ 3.6%, 8.2%]. However, the average precisions are slightly lower than the model Aeff, and the average recalls and the average F1s are slightly lower than the model Ares and Aeff. although they are obviously higher than the model B groups.

As the whole, it is expected to improve the accuracy of the AS prediction by constructing the model Ares and the model B4, and then, taking the ensemble of these two models, although it seems not to be necessary to take the ensemble of them in the view of the F1.

### 4.4 FFP and AIR

Several examples of the images' FFPs and AIRs extracted based on the model Ares and the model B4 are shown in Fig. 8. The images are rated to score 7, score 4, and score 2, respectively. Moreover, the image a and c are assigned to score 7 and 2 by either the Ares or the B4, respectively, while the image b is assigned to score 3 by the Ares, and to score 4 by the B4.

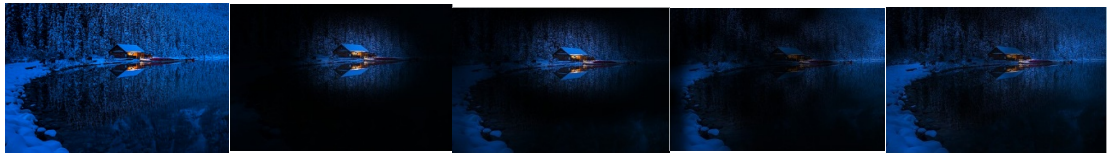

(a1) Original image　(a2) FFP by Ares　(a3) AIR by Ares　(a4) FFP by B4　(a5) AIR by B4

(a) Image a

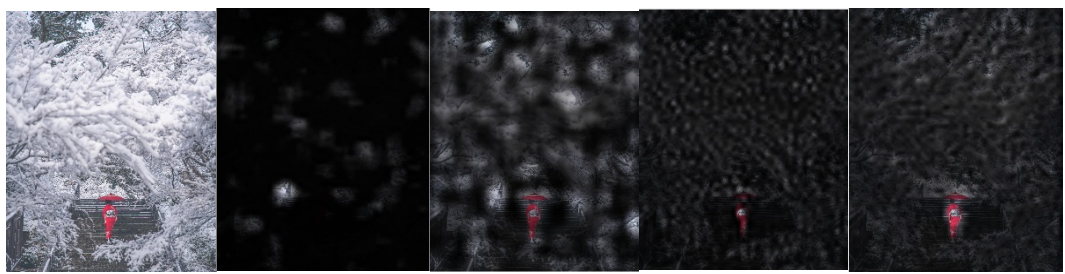

(b1) Original image    (b2) FFP by Ares    (b3) AIR by Ares    (b4) FFP by B4    (b5) AIR by B4

(b) Image b

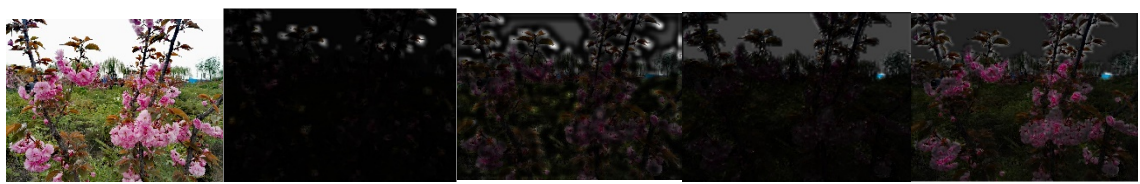

(c1) Original image    (c2) FFP by Ares    (c3) AIR by Ares    (c4) FFP by B4    (c5) AIR by B4

(c) Image c

Fig. 8 Several examples of the images' FFPs and AIRs

Similar with the Fig. 8, by observing the results of about 300 images, we can see that the FFPs and the AIRs extracted by the Ares are the objects in the images, while those extracted by the B4 seem to reflect the latent layouts of the objects. It seems that the images are assigned to the highly rated AS classes, such as the image a, if the FFPs and the AIRs meet the photography composition principles; the images are assigned to the lowly rated AS classes, such as the image c, if the FFPs and the AIRs don't meet the photography composition principles; the images are assigned to the mediately rated AS classes, such as the image b, if the FFPs and the AIRs appear the mediocre layouts. Whether or not the correct predictions are made, the above observations are similar. It is indicated that the models trained on XiheAA seem to learn the latent photography principles, but it cannot be said that they learned the aesthetic sense.

## 5  Conclusion

In this paper, we proposed a framework of constructing two types of IAA models with different CNN architectures and improving the performance of the image's AS prediction by the ensemble, and analyzing the effectiveness of the proposed methods on the XiheAA dataset and the CUHK-PQ public dataset. Moreover, it is found that the AS classification models trained on XiheAA seem to learn the latent photography composition principles by analyzing the FFPs and AIRs of the models to the images, although it cannot be said that they learn the aesthetic sense. Oh the other hand, although the precision, the recall, and the F1 of the AS prediction can't be said to be satisfied, it is sure that the proposed framework for the AS prediction is effective. The performance of the AS prediction should be improved, if more samples with higher rates or lower rates are collected to train the AS classification models.

## Acknowledgments

This work was partly supported by JSPS KAKENHI Grant Numbers 22K12095.